\documentclass{article}

\usepackage{PRIMEarxiv}

\usepackage[utf8]{inputenc} 
\usepackage[T1]{fontenc}    
\usepackage{hyperref}       
\usepackage{url}            
\usepackage{booktabs}       
\usepackage{amsfonts}       
\usepackage{nicefrac}       
\usepackage{microtype}      
\usepackage{lipsum}
\usepackage{fancyhdr}       
\usepackage{graphicx}       
\graphicspath{{media/}}     
\usepackage[percent]{overpic}
\usepackage{amsmath}
\usepackage{multirow}
\usepackage{makecell}

\newcommand{\GenSR}{G_{S \rightarrow R}}
\newcommand{\GenRS}{G_{R \rightarrow S}}

\title{Modified CycleGAN for the synthesization of samples for wheat head segmentation

}

\author{
  Jaden Myers \\
  Department of Computer Science \\
  University of Calgary \\
  Calgary, AB, Canada\\
  \texttt{jaden.myers@ucalgary.ca} \\
   \And
  Keyhan Najafian \\
  Department of Computer Science \\
  University of Saskatchewan \\
  Saskatoon, SK, Canada\\
  \texttt{keyhan.najafian@usask.ca} \\
  \AND
  Farhad Maleki \\
  Department of Computer Science \\
  University of Calgary \\
  Calgary, AB, Canada\\
  \texttt{farhad.maleki1@ucalgary.ca} \\
  \And
  Katie Ovens \thanks{Corresponding Author}\\
  Department of Computer Science \\
  University of Calgary \\
  Calgary, AB, Canada\\
  \texttt{katie.ovens@ucalgary.ca} \\
}

\begin{document}
\maketitle

\begin{abstract}
Deep learning models have been used for a variety of image processing tasks. However, most of these models are developed through supervised learning approaches, which rely heavily on the availability of large-scale annotated datasets. Developing such datasets is tedious and expensive. In the absence of an annotated dataset, synthetic data can be used for model development; however, due to the substantial differences between simulated and real data, a phenomenon referred to as domain gap, the resulting models often underperform when applied to real data. In this research, we aim to address this challenge by first computationally simulating a large-scale annotated dataset and then using a generative adversarial network (GAN) to fill the gap between simulated and real images. This approach results in a synthetic dataset that can be effectively utilized to train a deep-learning model. Using this approach, we developed a realistic annotated synthetic dataset for wheat head segmentation. This dataset was then used to develop a deep-learning model for semantic segmentation. The resulting model achieved a Dice score of 83.4\% on an internal dataset and Dice scores of 79.6\% and 83.6\% on two external Global Wheat Head Detection datasets. While we proposed this approach in the context of wheat head segmentation, it can be generalized to other crop types or, more broadly, to images with dense, repeated patterns such as those found in cellular imagery.
\end{abstract}


\section{Introduction}
Deep learning models have shown great potential for semantic image segmentation~\cite{SemanticSegmentationSurvey, DBLP:journals/corr/LongSD14}, where a label is assigned to each pixel representing its semantics. Most semantic segmentation models are developed using supervised approaches that rely on large amounts of annotated data. However, gathering large amounts of annotated data needed to train these models is often expensive, time-consuming~\cite{Richter_2016_ECCV}, and potentially requires highly specialized expertise to provide the image annotation. The need for a large, annotated data set can be alleviated by using synthetic annotated data. However, a model trained on simple synthetic data is unlikely to perform well when applied to real-world data. Therefore, effective integration of synthetic data into segmentation tasks requires overcoming the inherent distribution shift between synthesized and real images, thereby demanding domain adaptation strategies to bridge this gap.\par
Segmentation has been applied to annotate different plant components to detect abnormalities in crops such as lodging~\cite{mardanisamani2019crop}, and important phenotypic plant traits in precision agriculture, such as organ size~\cite{ImageSegmentationCanopyCover}, organ health~\cite{XIONG2020105712}, response to biotic~\cite{BioticStress} and abiotic stress~\cite{AbioticStress}. Large annotated datasets, such as the Global Wheat Head Detection (GWHD) dataset~\cite{david2021global}, have enabled the development of new supervised deep learning-based methods for plant phenotyping from field images. Since the pixel-level annotation required for segmentation tasks is substantially more demanding due to the time and effort invested, most of these datasets are focused on classification or object detection tasks and rarely on segmentation tasks.\par
Some recently proposed solutions for wheat heat segmentation aim to accomplish this task with little manual annotation~\cite{fourati2021wheat,najafian2021semi,najafian2023semi}. In these cases, synthesizing new samples to improve model performance for segmentation tasks is essential, but has the added challenge of distribution shift between the synthesized and real images and requires domain adaptation steps to gradually bridge the domain gap.  
\par
Fourati et al. utilized Faster R-CNN and EfficientDet models trained on the GWHD dataset to develop a wheat head detection method~\cite{fourati2021wheat}. They used semi-supervised techniques such as pseudo-labeling, test time augmentation, multi-scale ensemble, and bootstrap aggregation to improve model performance, and achieved a mean average precision of 0.74. Najafian et al. generated a synthesized dataset using a cut-and-paste approach and trained a YOLO architecture for wheat head detection~\cite{najafian2021semi}. Fine-tuning the model with the GWHD dataset resulted in a mean average precision of 0.82. However, their model showed lower performance when trained solely on the synthesized dataset (Precision: 0.318, Recall: 0.130).\par
Najafian et al. proposed a semi-self-supervised approach for semantic segmentation, utilizing both computationally and manually annotated images. They synthesized an image dataset for semantic segmentation models using a wheat field video clip, a few annotated images, and background scenes~\cite{najafian2023semi}. They trained a customized U-Net model with this synthesized dataset and applied domain adaptation steps to address the domain gap between synthesized and real images, and achieved a Dice score of 0.89 and 0.73 on their internal and external test sets, respectively. \par
This approach does come with some caveats. First, the distinction between the wheat heads in the synthesized images is highly contrasted with the background images where the wheat heads were overlaid. Therefore, this makes the problem of wheat head segmentation easier in the synthesized images used to train the model compared to actual images of wheat. This also means that the method is not easily extensible to other application areas beyond wheat head segmentation. Further, the initial synthesization of samples to train segmentation models to detect wheat heads does not reflect the realistic growth patterns of these crops. \par
We circumvent these pitfalls by training a model for the segmentation of wheat heads with a modified generative adversarial network (GAN) approach to generating the synthetic data set. A GAN traditionally consists of two subnetworks: a generative network and a discriminative network~\cite{goodfellow2014generative}. The role of the generative network is to generate realistic samples, while the discriminator model takes the generative network’s output as input along with real-world data and provides a binary classification output; true if the input is real or false if it is generated. The parameters of the generative model are adjusted according to the classifications made by the discriminative model and a loss function. The parameters of the discriminative model are adjusted in a similar way based on its classification of the data. The generative and discriminative models will be pitted against each other in a zero-sum game until the discriminative model is no longer able to consistently tell the difference between the synthetic and real data. This process results in a generative network that is able to generate images that closely resemble real-world images.\par
Cycle-Consistent Generative Adversarial Networks (CycleGANs), are a popular GAN-based deep learning framework focused on the task of unpaired image-to-image translation~\cite{zhu2017unpaired}. For example, they can be used to transform images of a crop field captured under different lighting conditions or with different camera types. This ability to translate images from a source domain to a target domain without the need for paired images allows for data augmentation, domain adaptation, and image enhancement in agriculture applications. The fundamental idea behind CycleGAN is to learn mappings between two different domains without the need for paired training data. Traditional methods for image translation, such as Pix2Pix~\cite{Pix2pix}, require paired examples of corresponding images in the source and target domains. However, acquiring such paired data can be challenging and labor-intensive or impractical in many cases. CycleGAN addresses this limitation by using unpaired data for training, making it more flexible and widely applicable. In agriculture, CycleGANs have been utilized for crop disease detection~\cite{tian2019detection}, plant phenotyping~\cite{li2023self}, and crop yield prediction~\cite{panda2010application}. \par
The limitation of simply using CycleGAN for the purpose of developing segmentation models with accurate annotation is that having an exact match between the generated image and the segmentation itself will not be enforced. A potential consequence of this is that some wheat heads generated do not have a corresponding segmentation, leading to an inaccurate data annotation that will negatively affect downstream models trained on these images. Therefore, we propose to adapt CycleGAN by adding this enforcement step to maintain consistency between generated images and their corresponding annotation.\par
The paper is structured as follows: In Section~\ref{SEC:MaterialsMethods}, we provide an overview of the modified CycleGAN architecture employed to generate synthetic training samples for wheat head segmentation. Section~\ref{SEC:Results} presents the results and evaluation of our proposed model. Finally, Section~\ref{SEC:Discussion} discusses the findings, and Section~\ref{SEC:Conclusion} provides the concluding remarks.
\section{Materials and Methods}
\label{SEC:MaterialsMethods}
\subsection{Datasets}
\begin{figure*}[!tbph]
    \centering
    \includegraphics[width=\textwidth]{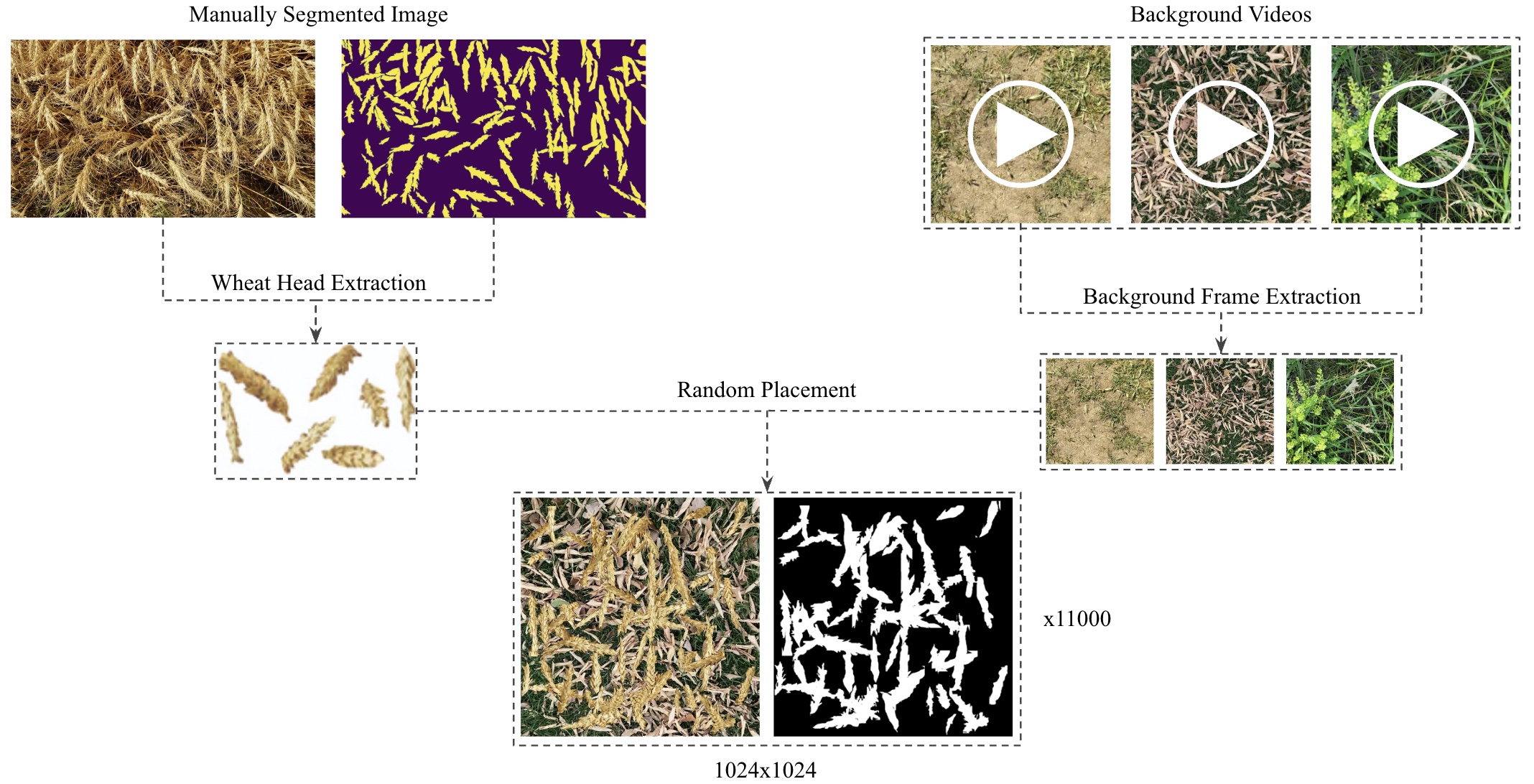}
    \caption{A visualization of the pipeline used to generate synthetic images. Wheat head cutouts are extracted from a manually annotated real wheat image and background frames are extracted from the background videos. The wheat heads are then randomly overlaid onto background frames to generate a wheat head image and a semantic segmentation mask.}
    \label{fig:synthesization}
\end{figure*}

The data used in this study can be accessed at  \href{https://www.cs.usask.ca/ftp/pub/whs/}{https://www.cs.usask.ca/ftp/pub/whs/}. All the data used to train and validate our models comes from a video clip of a wheat field, hereafter referred to as $W$, and 11 videos of background scenes captured using Samsung cameras with resolutions of 12 and 48 Megapixels. With only a single frame from $W$ that has been manually annotated and frames from the background videos, we generated a synthetic dataset $S$ using the cut-and-paste method proposed by Najafian et al.~\cite{najafian2023semi}. Figure~\ref{fig:synthesization} illustrates this approach. The dataset $S$ consists of 11,000 synthetic images and corresponding semantic segmentation labels. The rest of the frames in $W$, aside from 100 frames which will be used for testing, become the dataset $R$ consisting of real wheat head images with no ground truth semantic segmentation labels. Our goal is to use the datasets $S$ and $R$ to generate a dataset $\hat{R}$ consisting of GAN-based synthetic images that more closely resemble the real images of $R$. $\hat{R}$ will consist of 11,000 synthetic images and ground truth semantic segmentation labels, 10,000 of which will be used for training and 1,000 for validation of a semantic segmentation model.\par
To evaluate the performance of the segmentation model trained on our dataset $\hat{R}$, we test its performance on 3 different evaluation data sets: 1) an internal dataset consisting of the 100 manually annotated images from $W$, 2) an external dataset consisting of 365 samples across 18 different domains from the GWHD dataset~\cite{david2021global} which was annotated by Najafian et al.~\cite{najafian2023semi} to evaluate their model, and 3) another external dataset consisting of samples from the UTokyo subset of the GWHD dataset~\cite{david2021global} which was annotated by Najafian et al.~\cite{najafian2023semi} to evaluate their model and allows us to compare the performance of our model with the model developed by Rawat et al.~\cite{UTokyo}. As a result of the internal test set consisting of images from the same video from which we generate the dataset $\hat{R}$, the performance of the model on the internal test set may not be a good indication of the models ability to generalize to new domains. 
Therefore, we use these two external test datasets---consisting of different varieties of wheat, in different fields, and at various stages of growth, none of which have been seen during training by the model.

\subsection{Model architecture}
\begin{figure*}[!tbph]
    \centering
    \includegraphics[width=\textwidth]{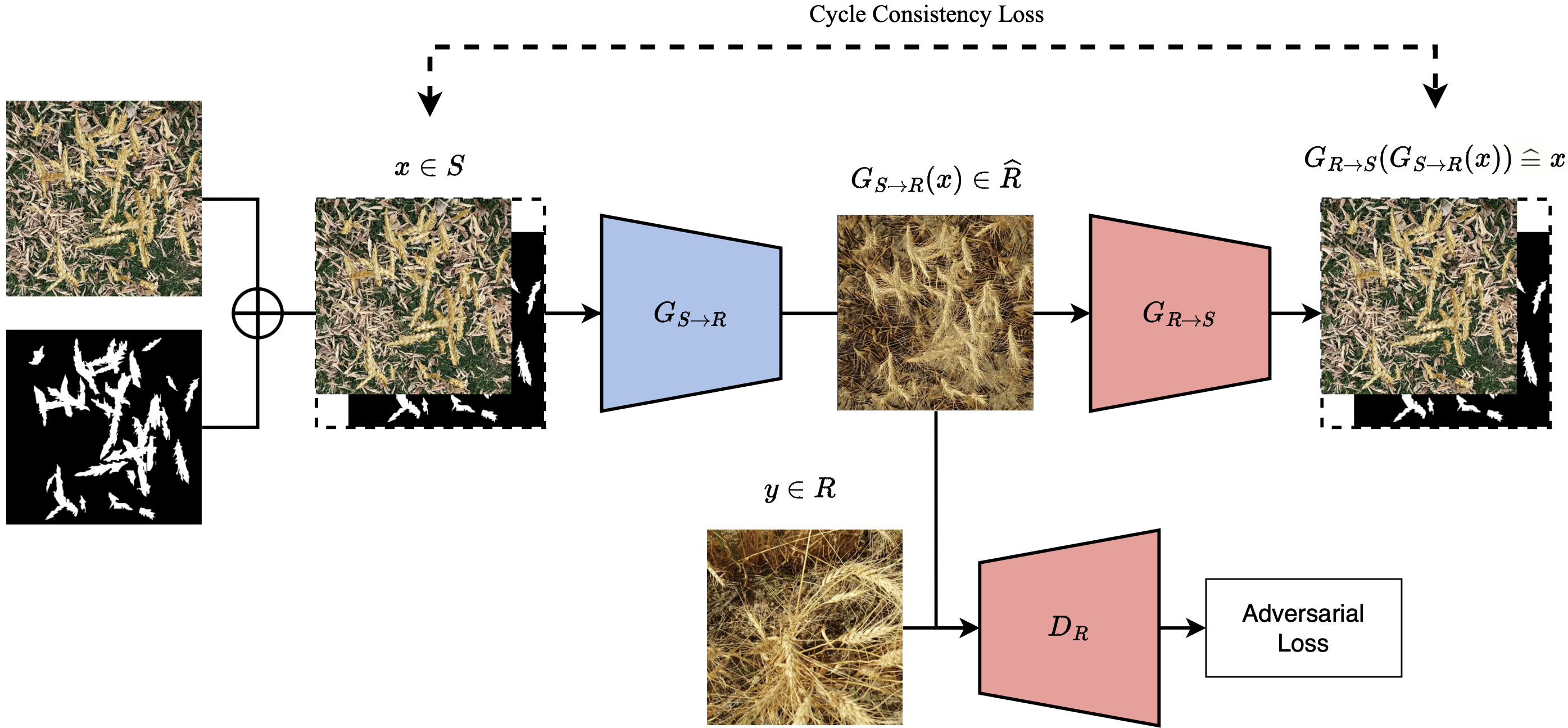}
    \caption{Diagram of the modified CycleGAN. The generator $\GenSR :S \rightarrow R$ takes as input synthetic images concatenated with their semantic segmentation masks $x \in S$ and outputs a corresponding real image $\GenSR(x) \in \hat{R}$. A cycle consistency loss is calculated between $x$ and $\GenRS(\GenSR(x))$. Not present in the diagram, a cycle consistency loss is also calculated in the opposite direction with real images $y \in R$ and $\GenSR(\GenRS(y))$, and the discriminator $D_{S}$ calculates an adversarial loss with $\GenRS(Y) \in S$ and $x$.}
    \label{fig:cycle}
\end{figure*}

The proposed model architecture is presented in Figure~\ref{fig:cycle}. This architecture is a modified CycleGAN~\cite{zhu2017unpaired} which uses the input and recreation of semantic segmentation masks to enforce the preservation of semantic information through the image translation. CycleGAN is an image translation framework that utilizes an adversarial training objective and a cycle consistency loss to achieve unpaired image-to-image translation. The enforced cycle consistency forces the generator to produce domain-translated images that still resemble the original images. The overall objective function combines adversarial losses from the discriminator and cycle consistency losses. By optimizing this objective function, the generator and discriminator networks can learn to perform unsupervised image translation without the need for paired training data. Although the unmodified CycleGAN focuses on the overall visual appearance, it does not explicitly impose constraints to enforce semantic consistency for the masked regions (wheat heads here) during the translation process. Therefore, we update the model architecture to impose constraints on the generated images. First, given an input image and its segmentation mask, the first generator produces an image in the style of the target domain without a segmentation mask. Next, the second generator takes the generated images and outputs the recreated image along with the recreated mask. The cycle consistency loss is then calculated by measuring the distance between the original image and the recreated image, as well as the original and recreated mask. This encourages the model to preserve semantic loss through image translation because the mask must be recreated at the end of the cycle.

For the segmentation model, we use the same modified U-Net~\cite{U-Net} which was utilized by Najafian et al.~\cite{najafian2023semi}. A binary cross-entropy loss function is used to train all segmentation models. For evaluation of segmentation performance, Dice and IoU are used.

\subsection{Pseudo Labelling}
\begin{figure*}[!tbph]
    \centering
    \includegraphics[width=\textwidth]{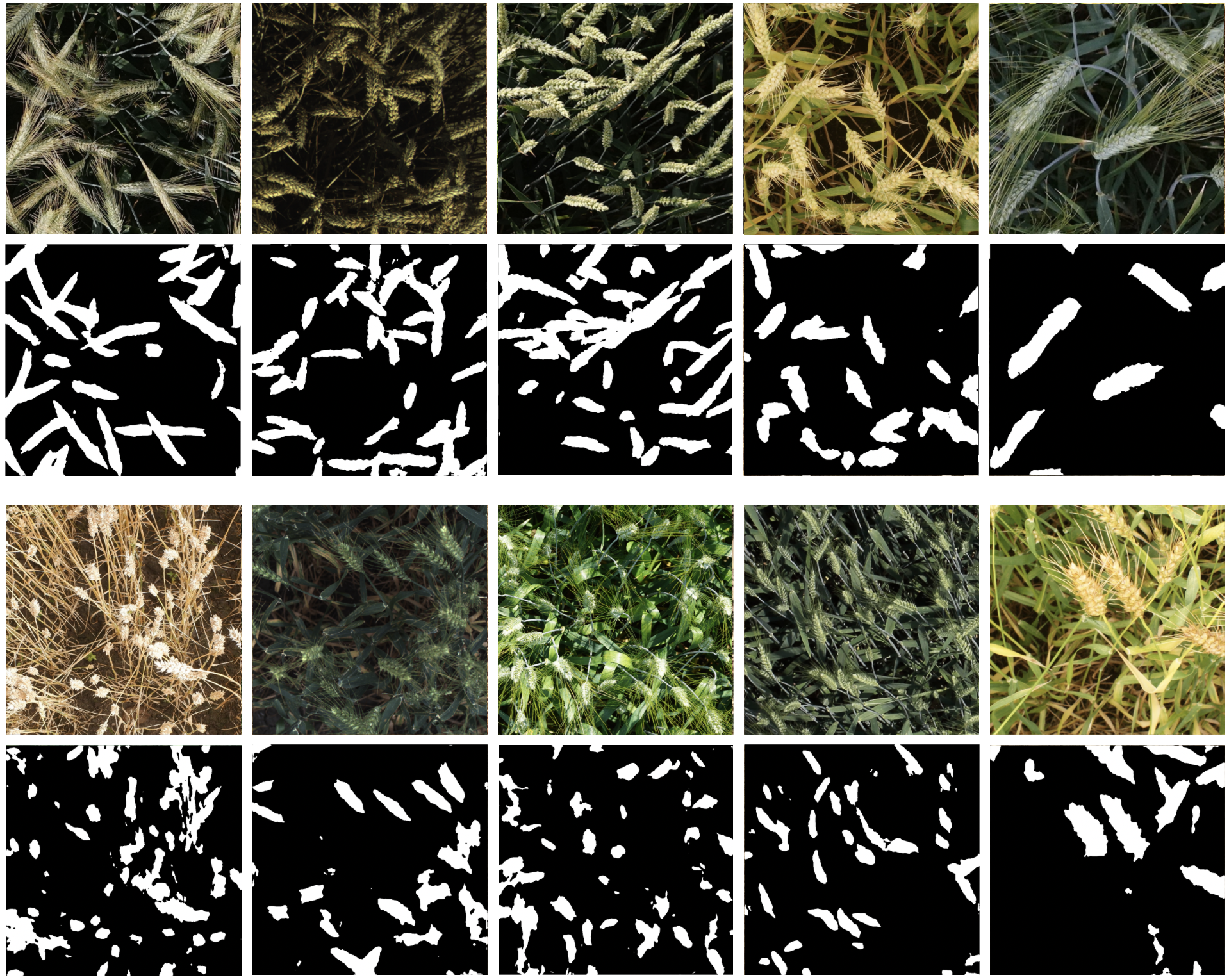}
    \caption{Examples of randomly selected GWHD images and the corresponding pseudo mask predictions. The top row consists of samples that were selected to be part of the dataset used to fine-tune the model. The bottom row consists of samples that were not selected.}
    \label{fig:pseudo_labels}
\end{figure*}

To enhance the performance of the model trained on the dataset $\hat{R}$, we introduce a domain adaptation step based on the pseudo label (PL) approach ~\cite{Pseudo-Label}. From the GWHD, we select 360 random images that are not part of either GWHD external evaluation datasets and pass them as input to the model trained on $\hat{R}$ to get the predicted segmentation masks. From these predictions, we select only the best predictions, which are then used to fine-tune the model. Choosing these predictions is somewhat of a subjective process, but there are criteria that can be used to determine the quality of a prediction. Figure~\ref{fig:pseudo_labels} presents examples of images that were selected and not selected. A prediction is of high quality and should be selected if it maximizes the area of the wheat head in the image being segmented out and minimizes the nonwheat head parts of the image being segmented out. A dataset of 99 pseudo-labeled images was compiled using the process.

\section{Results}

\label{SEC:Results}

\subsection{Synthesization of wheat head images}
\begin{figure*}[!tbph]
    \centering
    \includegraphics[width=\textwidth]{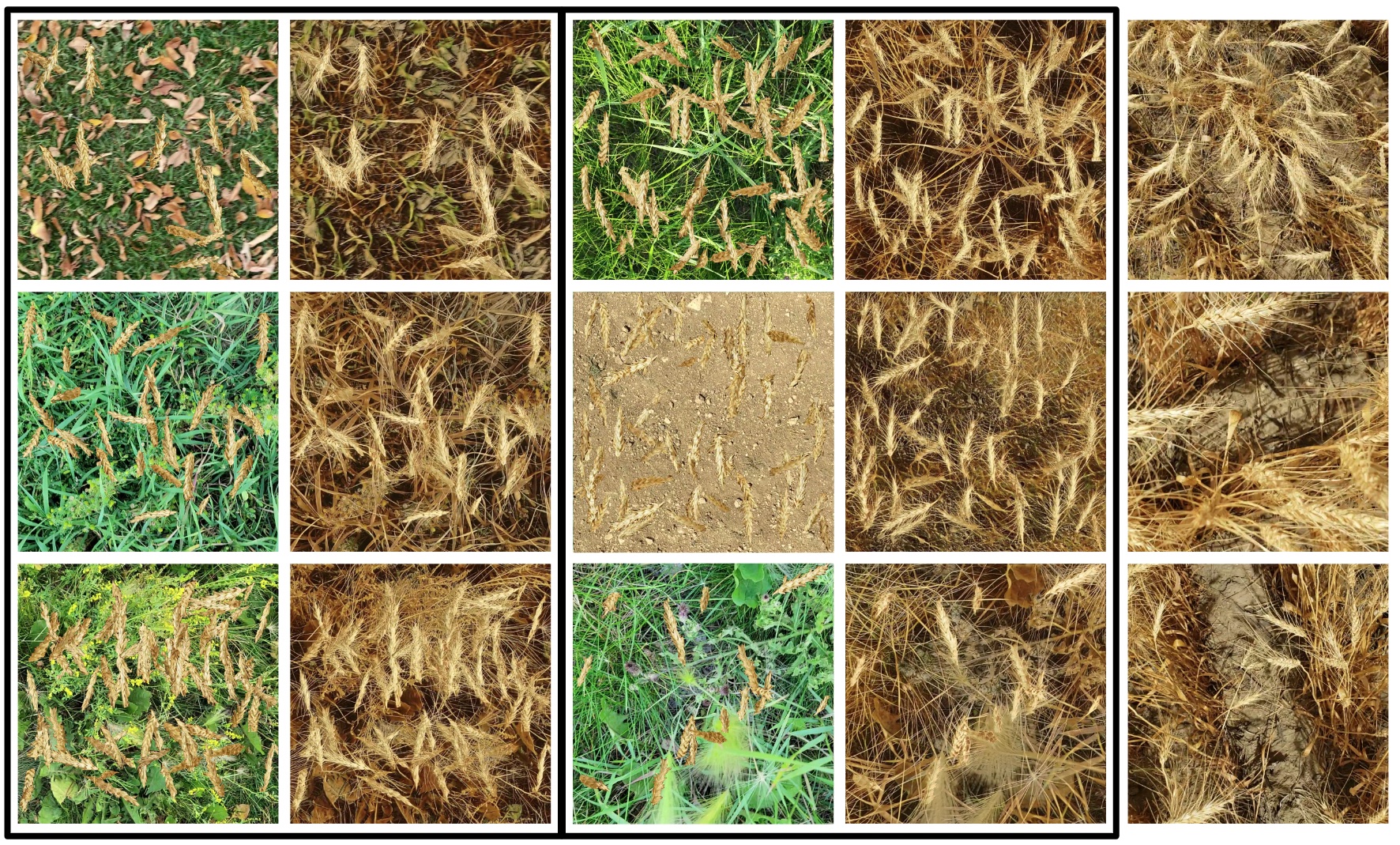}
    \caption{In the boxes are randomly selected synthetic images on the left and the corresponding outputs from our modified CycleGAN on the right. The images on the far right are randomly selected real wheat images for comparison.}
    \label{fig:synthesized}
\end{figure*}

\begin{figure}[!tbph]
    \centering
    \includegraphics[width=\textwidth]{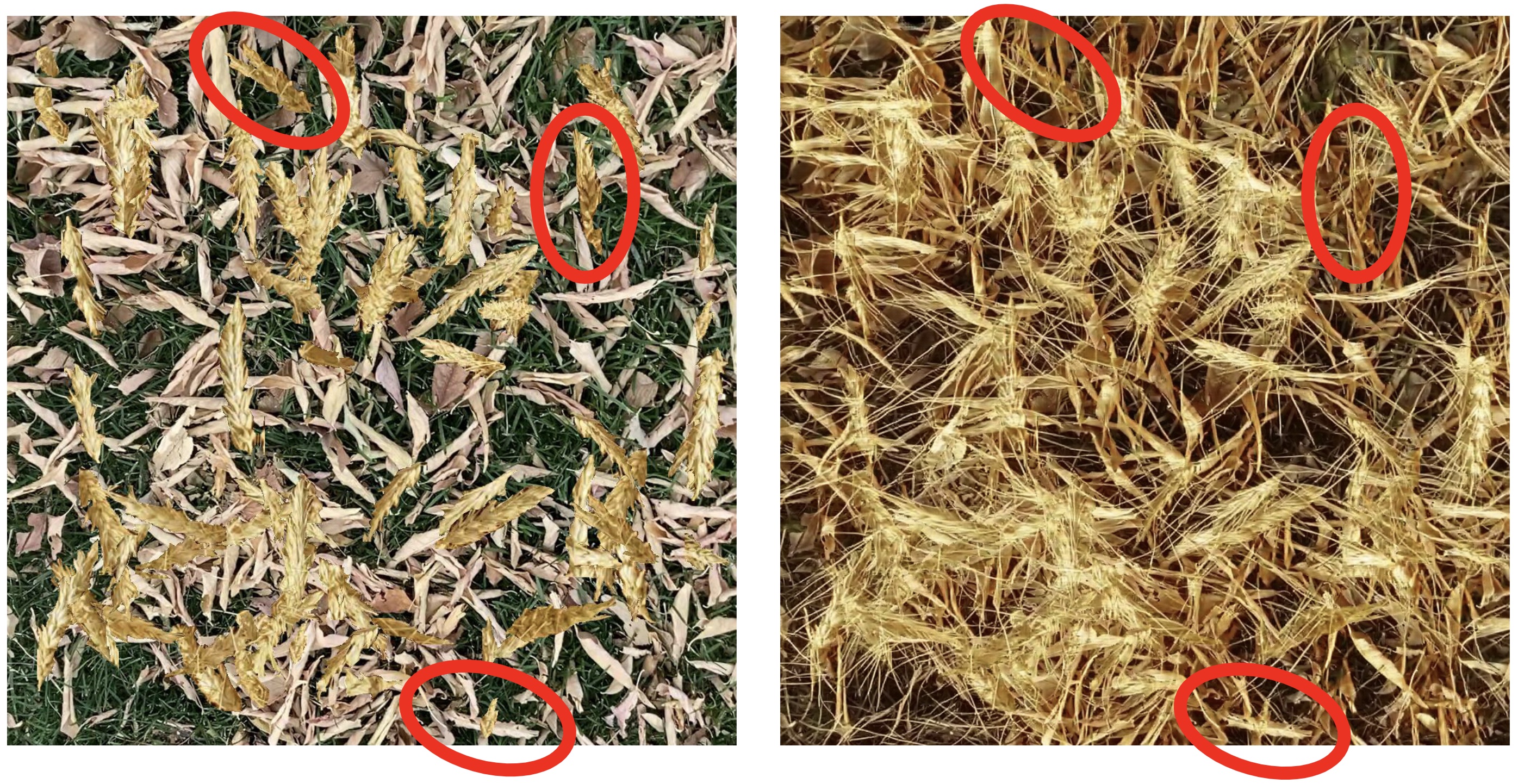}
    \caption{Synthetic wheat image and the corresponding output from an unmodified CycleGAN. The red circles highlight the flaws of the unmodified CycleGAN image translation.}
    \label{fig:regular-cycle}
\end{figure}

\begin{figure}[!tbph]
    \centering
    \includegraphics[width=0.8\textwidth]{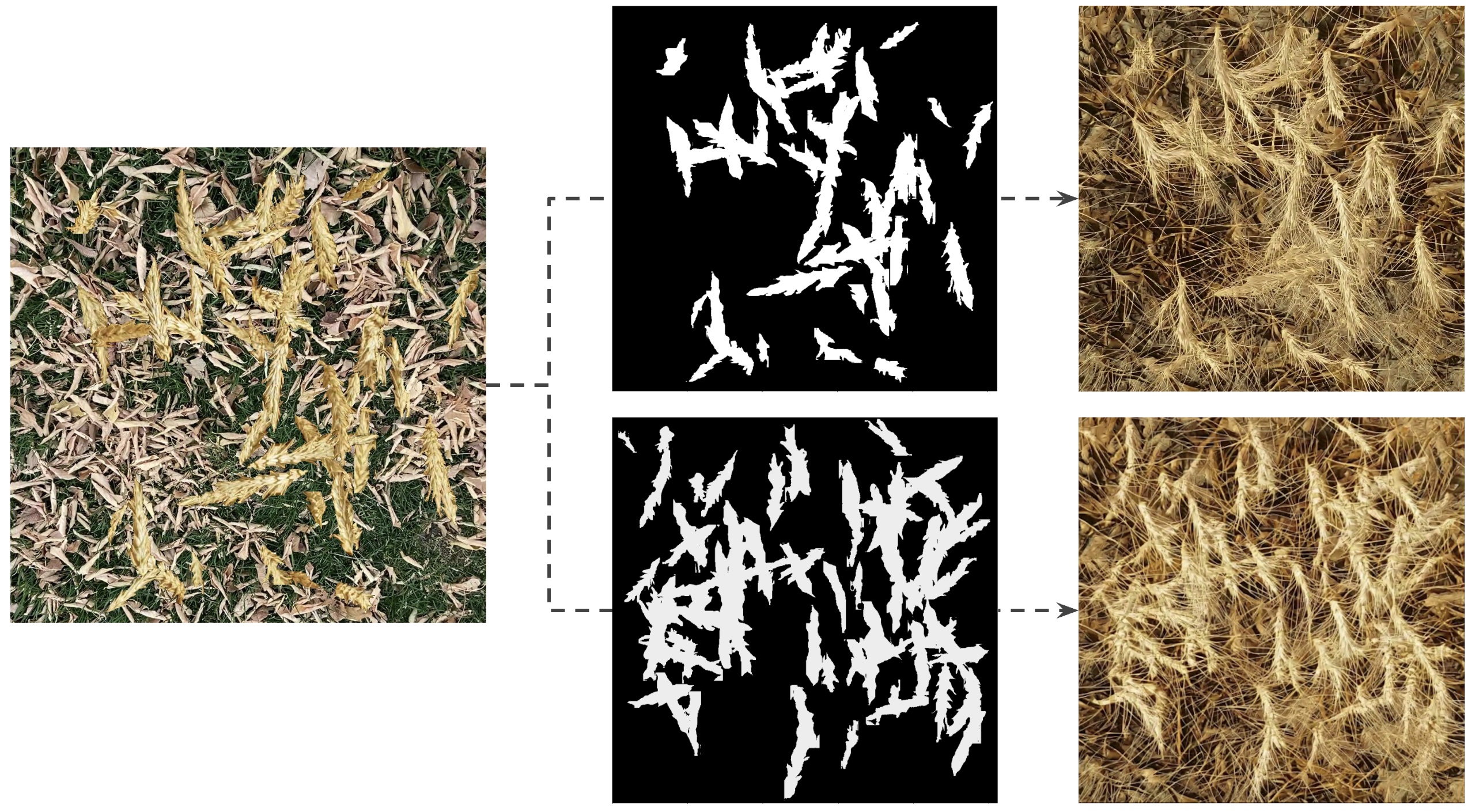}
    \caption{The same synthetic wheat image concatenated with two different semantic segmentation masks and the corresponding output from our modified CycleGAN.}
    \label{fig:mask}
\end{figure}

Figure ~\ref{fig:synthesized} shows randomly selected synthetic images from the dataset $S$, and the corresponding translated output from the modified CycleGAN in the dataset $\hat{R}$ with real wheat images for comparison. The translated synthetic images in $\hat{R}$ are visibly closer to matching the realistic images compared to the synthetic images from the dataset $S$. Although the translated images from the regular CycleGAN look realistic, the semantic features are not preserved through the translation as illustrated by Figure ~\ref{fig:regular-cycle}. Our CycleGAN with segmentation mask input learned to preserve the semantic features through translation as shown by Figure ~\ref{fig:mask} which illustrates the model's ability to place wheat heads corresponding to the input segmentation mask.

\subsection{Evaluation of wheat head segmentation model trained with generated wheat head images}
The performance of the segmentation models is outlined in Table~\ref{table:performance}. Compared to the model trained on the dataset $S$, the model trained on our CycleGAN augmented synthetic dataset $\hat{R}$ showed a significant increase of 0.1016 and 0.1200 in the Dice and IoU, respectively, when evaluated on the internal evaluation dataset. When evaluated on the external GWHD test set consisting of samples from 18 domains, the performance of the model trained on our dataset showed an even more substantial increase of $0.2098$ and $0.1654$ in the Dice and IoU score, respectively. When evaluated on the GWHD test set from UTokyo, the model trained on our data saw an increase of $0.2337$ and $0.2357$. Furthermore, after the model trained on the dataset $\hat{R}$ was fine-tuned using the PL approach, the model achieved an even greater increase in performance across all evaluation datasets.
\begin{table}[!tbhp]
\renewcommand{\arraystretch}{1.5} 
\caption{The performance of the models trained on the synthetic data sets.}
\label{table:performance}
\footnotesize
\begin{tabular}{|l|c|c|c|c|c|c|c|c|c|}
\hline
Evaluation Set & \multicolumn{3}{c|}{Internal}        & \multicolumn{3}{c|}{GWHD (18 Domains)} & \multicolumn{3}{c|}{GWHD (UTokyo)}   \\\hline
Training Set   & S      & $\hat{R}$ & $\hat{R}$ + PL & S       & $\hat{R}$  & $\hat{R}$ + PL & S      & $\hat{R}$ & $\hat{R}$ + PL \\\hline
Dice Score     & 0.7090  & 0.8106    & 0.834          & 0.3678  & 0.5776     & 0.7958         & 0.4067 & 0.6444    & 0.8356         \\\hline
IoU Score      & 0.5658 & 0.6858    & 0.7196         & 0.2742  & 0.4403     & 0.6874         & 0.2749 & 0.5106    & 0.7535        \\\hline
\end{tabular}

\end{table}
\section{Discussion}

\label{SEC:Discussion}
We proposed a method for computationally generating large amounts of robust labeled training data for the purpose of training a wheat head segmentation model. The proposed approach, utilizing only one manually annotated image, allowed us to generate a synthetic training dataset and train a model that has a much smaller domain gap to bridge when applied to real-life data. In many domains, such as medical imaging and precision agriculture, annotated training data can be hard to come by due to the expensive and labor-intensive task of manual annotation. Furthermore, in the medical domain, specialized expertise is often needed to carry out these tasks. As a result of our proposed method only using a single manually annotated image, our approach has the potential to be applied to such domains where other deep learning approaches would be limited by the lack of annotated data.

\subsection{Synthetic Image Generation}
In the previous approach, images were synthesized by inserting extracted wheat head cutouts onto background frames extracted from videos. While this approach gives us a way to computationally generate large amounts of synthetic training data, the resulting data does not resemble real data which is reflected in the poor performance of the segmentation models trained on the data when applied to real test data. The proposed approach adds a new step to the image synthesization pipeline. We do image-to-image translation using a modified CycleGAN to make the simple synthetic data resemble real wheat data. The resulting synthetic images visibly resemble real-life data substantially more than the simple synthetic data.\par

When using the CycleGAN image synthesization approach with an unmodified CycleGAN, the resulting data looks realistic. However, during the image translation process with the unmodified CycleGAN, semantic information was not preserved. This is reflected in the poor performance of the U-Net trained on unmodified CycleGAN data. Our modified CycleGAN with semantic mask input was able to generate real-looking images while also preserving the semantic information of the images.

\subsection{Limitations and future directions}
To reach high performance, our approach requires manual input in the form of careful curation of pseudo-labeled data for further model training. Manual input such as this is expensive can introduce bias into the system and is hard to reproduce. Thus, future projects should aim to improve the process for generating a synthetic training dataset so that a model can be trained for high performance without the need for manual input.

In our proposed modified CycleGAN, the model must recreate masks for the purpose of preserving semantic information during image translation. Utilizing a model for the generator that is better suited for creating semantic masks could result in better preservation of semantic information and thus lead to more robust training data. A generator architecture such as the U-Net generator proposed by Torbunov et al.~\cite{torbunov2022uvcgan} could be used to achieve this. However, this approach could lead to a trade off between the quality of generated images and semantic preservation as the U-Net generator has been shown to under-perform in the image to image translation task compared to the generator architecture we used in this study.\par

To train the CycleGAN for synthetic to real image translation, we used real images extracted from a video clip of a single wheat field. Since all real training images were extracted from the same video clip, there is minimal variation within the real training data. Utilizing multiple images and/or videos taken from multiple wheat fields would increase the variation within the real data and could potentially lead to the data generated by the CycleGAN being more robust as training data.\par

In the first step of image synthesization, wheat heads are randomly overlaid on the background frames. This does not perfectly emulate the layout of wheat heads in real images, and our CycleGAN approach does nothing to make the layout of the images more resemble that of real images. Creating a pipeline that uses density maps to resemble the rows and orientations of wheat heads in real wheat images instead of complete random placement could further decrease the domain gap between the synthetic and real wheat data.\par

\section{Conclusions}
\label{SEC:Conclusion}
This study proposed an approach for the synthesization of training data for wheat head segmentation. Making use of a modified CycleGAN for the synthesization of images allowed us to computationally generate large amounts of robust synthetic data that resembles real data. Our findings highlight the potential of utilizing synthetic data for deep learning applications in domains where annotated data is limited and costly to produce.


\bibliographystyle{unsrt}  
\bibliography{references}

\begin{thebibliography}{10}

\bibitem{SemanticSegmentationSurvey}
Shijie Hao, Yuan Zhou, and Yanrong Guo.
\newblock A brief survey on semantic segmentation with deep learning.
\newblock {\em Neurocomputing}, 406:302--321, 2020.

\bibitem{DBLP:journals/corr/LongSD14}
Jonathan Long, Evan Shelhamer, and Trevor Darrell.
\newblock Fully convolutional networks for semantic segmentation.
\newblock {\em CoRR}, abs/1411.4038, 2014.

\bibitem{Richter_2016_ECCV}
Stephan~R. Richter, Vibhav Vineet, Stefan Roth, and Vladlen Koltun.
\newblock Playing for data: {G}round truth from computer games.
\newblock In Bastian Leibe, Jiri Matas, Nicu Sebe, and Max Welling, editors, {\em European Conference on Computer Vision (ECCV)}, volume 9906 of {\em LNCS}, pages 102--118. Springer International Publishing, 2016.

\bibitem{mardanisamani2019crop}
Sara Mardanisamani, Farhad Maleki, Sara Hosseinzadeh~Kassani, Sajith Rajapaksa, Hema Duddu, Menglu Wang, Steve Shirtliffe, Seungbum Ryu, Anique Josuttes, Ti~Zhang, et~al.
\newblock Crop lodging prediction from {UAV}-acquired images of wheat and canola using a {DCNN} augmented with handcrafted texture features.
\newblock In {\em Proceedings of the {IEEE/CVF} conference on computer vision and pattern recognition workshops}, 2019.

\bibitem{ImageSegmentationCanopyCover}
Biao Jia, Haibing He, Fuyu Ma, Ming Diao, Guiying Jiang, Zhong Zheng, Jin Cui, and Hua Fan.
\newblock Use of a digital camera to monitor the growth and nitrogen status of cotton.
\newblock {\em TheScientificWorldJournal}, 2014:602647, 02 2014.

\bibitem{XIONG2020105712}
Yonghua Xiong, Longfei Liang, Lin Wang, Jinhua She, and Min Wu.
\newblock Identification of cash crop diseases using automatic image segmentation algorithm and deep learning with expanded dataset.
\newblock {\em Computers and Electronics in Agriculture}, 177:105712, 2020.

\bibitem{BioticStress}
Jan Behmann, Anne-Katrin Mahlein, T.~Rumpf, Christoph Römer, and Lutz Plümer.
\newblock A review of advanced machine learning methods for the detection of biotic stress in precision crop protection.
\newblock {\em Precision Agriculture}, 16:239--260, 06 2015.

\bibitem{AbioticStress}
Nadia Al-Tamimi, Patrick Langan, Villő Bernád, Jason Walsh, Eleni Mangina, and Sonia Negrao.
\newblock Capturing crop adaptation to abiotic stress using image-based technologies.
\newblock {\em Open Biology}, 12, 06 2022.

\bibitem{david2021global}
Etienne David, Mario Serouart, Daniel Smith, Simon Madec, Kaaviya Velumani, Shouyang Liu, Xu~Wang, Francisco Pinto, Shahameh Shafiee, Izzat~SA Tahir, et~al.
\newblock Global wheat head detection 2021: An improved dataset for benchmarking wheat head detection methods.
\newblock {\em Plant Phenomics}, 2021.

\bibitem{fourati2021wheat}
Fares Fourati, Wided~Souidene Mseddi, and Rabah Attia.
\newblock Wheat head detection using deep, semi-supervised and ensemble learning.
\newblock {\em Canadian Journal of Remote Sensing}, 47(2):198--208, 2021.

\bibitem{najafian2021semi}
Keyhan Najafian, Alireza Ghanbari, Ian Stavness, Lingling Jin, Gholam~Hassan Shirdel, and Farhad Maleki.
\newblock A {Semi-Self-Supervised} learning approach for wheat head detection using extremely small number of labeled samples.
\newblock In {\em Proceedings of the {IEEE/CVF} International Conference on Computer Vision}, pages 1342--1351, 2021.

\bibitem{najafian2023semi}
Keyhan Najafian, Alireza Ghanbari, Mahdi Sabet~Kish, Mark Eramian, Gholam~Hassan Shirdel, Ian Stavness, Lingling Jin, and Farhad Maleki.
\newblock {Semi-Self-Supervised} learning for semantic segmentation in images with dense patterns.
\newblock {\em Plant Phenomics}, 5:0025, 2023.

\bibitem{goodfellow2014generative}
Ian Goodfellow, Jean Pouget-Abadie, Mehdi Mirza, Bing Xu, David Warde-Farley, Sherjil Ozair, Aaron Courville, and Yoshua Bengio.
\newblock Generative adversarial nets.
\newblock {\em Advances in neural information processing systems}, 27, 2014.

\bibitem{zhu2017unpaired}
Jun-Yan Zhu, Taesung Park, Phillip Isola, and Alexei~A Efros.
\newblock Unpaired image-to-image translation using cycle-consistent adversarial networks.
\newblock In {\em Proceedings of the {IEEE} international conference on computer vision}, pages 2223--2232, 2017.

\bibitem{Pix2pix}
Phillip Isola, Jun-Yan Zhu, Tinghui Zhou, and Alexei~A. Efros.
\newblock Image-to-image translation with conditional adversarial networks.
\newblock In {\em 2017 IEEE Conference on Computer Vision and Pattern Recognition (CVPR)}, pages 5967--5976, 2017.

\bibitem{tian2019detection}
Yunong Tian, Guodong Yang, Zhe Wang, En~Li, Zize Liang, et~al.
\newblock Detection of apple lesions in orchards based on deep learning methods of cyclegan and yolov3-dense.
\newblock {\em Journal of Sensors}, 2019, 2019.

\bibitem{li2023self}
Yinglun Li, Xiaohai Zhan, Shouyang Liu, Hao Lu, Ruibo Jiang, Wei Guo, Scott Chapman, Yufeng Ge, Benoit Solan, Yanfeng Ding, et~al.
\newblock Self-supervised plant phenotyping by combining domain adaptation with {3D} plant model simulations: application to wheat leaf counting at seedling stage.
\newblock {\em Plant Phenomics}, 5:0041, 2023.

\bibitem{panda2010application}
Sudhanshu~Sekhar Panda, Daniel~P Ames, and Suranjan Panigrahi.
\newblock Application of vegetation indices for agricultural crop yield prediction using neural network techniques.
\newblock {\em Remote Sensing}, 2(3):673--696, 2010.

\bibitem{UTokyo}
Shivangana Rawat, Akshay L~Chandra, Sai~Vikas Desai, Vineeth Balasubramanian, Seishi Ninomiya, and Wei Guo.
\newblock How useful is image-based active learning for plant organ segmentation?
\newblock {\em Plant Phenomics}, 2022:1--11, 02 2022.

\bibitem{U-Net}
Olaf Ronneberger, Philipp Fischer, and Thomas Brox.
\newblock U-net: Convolutional networks for biomedical image segmentation.
\newblock {\em CoRR}, abs/1505.04597, 2015.

\bibitem{Pseudo-Label}
Dong-Hyun Lee.
\newblock Pseudo-label : The simple and efficient semi-supervised learning method for deep neural networks.
\newblock {\em ICML 2013 Workshop : Challenges in Representation Learning (WREPL)}, 07 2013.

\bibitem{torbunov2022uvcgan}
Dmitrii Torbunov, Yi~Huang, Haiwang Yu, Jin Huang, Shinjae Yoo, Meifeng Lin, Brett Viren, and Yihui Ren.
\newblock Uvcgan: Unet vision transformer cycle-consistent gan for unpaired image-to-image translation, 2022.

\end{thebibliography}

\end{document}